# EPAN: Robust Pedestrian Re-Identification via Enhanced Alignment Network for IoT Surveillance

Zhiyang ia*, Hongyan Cui*#, *Senior Member, IEEE*, Ge Gao*, Bo Li*, Minjie Zhang*, Zishuo Gao*, Huiwen Huang* and Caisheng Zhuo*

***Abstract*—Person re-identification (ReID) plays a pivotal role in computer vision, particularly in surveillance and security applications within IoT-enabled smart environments. This study introduces the Enhanced Pedestrian Alignment Network (EPAN), tailored for robust ReID across diverse IoT surveillance conditions. EPAN employs a dual-branch architecture to mitigate the impact of perspective and environmental changes, extracting alignment information under varying scales and viewpoints. Here, we demonstrate EPAN's strong feature extraction capabilities, achieving outstanding performance on the Inspection-Personnel dataset with a Rank-1 accuracy of 90.09% and a mean Average Precision (mAP) of 78.82%. This highlights EPAN's potential for real-world IoT applications, enabling effective and reliable person ReID across diverse cameras in surveillance and security systems. The code and data are available at: https://github.com/ggboy2580/EPAN**

***Index Terms*—Person re-identification, Affine network, Camera angle variations, Deep learning.**

## I. INTRODUCTION

THE goal of pedestrian re-identification (ReID) in inspection scenarios involving maintenance personnel is to enable efficient and accurate retrieval of target individuals from vast and complex databases, a critical subtask in IoT-enabled surveillance systems[1],[2],[3],[4].This challenge, a subdomain of computer vision focusing on image retrieval, holds significant value in industrial IoT applications, such as automated employee attendance monitoring and real-time suspicious individual tracking[5],[6],[7],[8]. These applications not only enhance the vertical integration of artificial intelligence (AI) technologies in industrial domains but also contribute to the broader development of AI-driven IoT systems[9],[10],[11],[12].

Traditional computer vision techniques for pedestrian feature matching have struggled with limitations in accuracy and robustness, confining their utility to controlled experimental settings [13],[14],[15],[16]. And the advent of large-scale datasets and advancements in deep learning have significantly propelled pedestrian ReID technologies forward [17],[18],[19],[20],[21]. Modern network architectures like MobileNet [22], OSNet [23], ShuffleNet [24], ResNet [25], and VggNet [26] have demonstrated strong capabilities in feature extraction. Recently, many teams have been researching specific pedestrian re identification problems [27],[28],[29],[30] such as video pedestrian re identification technology, domain migration pedestrian re identification[31], occluded pedestrian re identification technology[32], unmanned aerial vehicle perspective pedestrian re identification technology [33], extracting visual features and spatiotemporal information through multi view correlation[34], proposing robust dual embedding methods to improve cross modal ReID's ability to cope with noise[35], proposing privacy preserving anonymization methods to generate reversible full body anonymous images[36], and combining texture and contour features to improve the performance of video based ReID[37]. Nonetheless, specific industrial IoT app lications require tailored modifications to these architectures to address real-world challenges. The following section systematically clarifies the core pain points of current ReID technologies in industrial IoT surveillance scenarios and the research motivation of this study.

### *A. Research Motivation and Problem Focus*

Industrial IoT surveillance scenarios (e.g., power infrastructure inspection, equipment room personnel management) differ fundamentally from general ReID scenarios (e.g., public squares, shopping malls), leading to three core bottlenecks that limit the practical application of existing ReID technologies:

**Post-Detection Image Misalignment:** Surveillance cameras in inspection scenarios are typically installed at low angles (below human height)—a design constraint to monitor equipment operations. This results in two critical issues for pedestrian detection: (1) Excessive background noise: Detection boxes often include irrelevant objects such as equipment cabinets, pipelines, and tool racks, which "pollute" feature extraction (e.g., traditional models may mistake cabinet textures for pedestrian clothing features); (2) Partial occlusion: Pedestrian body parts (e.g., heads, legs) are frequently blocked by inspection equipment, leading to incomplete feature representation. Traditional ReID models adopt a "detection-then-recognition" pipeline, which cannot dynamically correct misaligned detection results, directly reducing feature extraction accuracy [45],[46].

This work was supported by National Natural Science Foundation of China(62171049), China University of Petroleum (Beijing) Karamay Campus introduction of talents and launch of scientific research projects(XQZX20240010), and China Tower Corporation Limited IT System 2023 Package Software Project - AI Algorithm and Services (23M01ZBZB011000017). *(Zhiyang jia, Hongyan Cui, Ge Gao, Bo Li are co-first authors. ).*

Hongyan Cui is with Beijing University of Posts and Telecommunications, Beijing, China, 100876. (e-mail: cuihy@bupt.edu.cn).

Zhiyang Jia, Ge Gao, Minjie Zhang , Zishuo Gao, Huiwen Huang and Caisheng Zhuo are with China University of Petroleum (Beijing) Karamay Campus, Beijing, China,834000. (e-mail: 2018592015@cupk.edu.cn; gaoge@st.cupk.edu.cn; 2023591304@cupk.edu.cn; 2023016002@st.cupk.edu.cn; 2023216711@st.cupk.edu.cn; 2023216709@st.cupk.edu.cn).

Bo Li is with Information Technology Research Institute of China Tower, China, 100089. (libo88872@chinatowercom.cn)

**Insufficient Environmental Robustness:**Inspection environments are characterized by closed spaces, abrupt illumination changes, and homogeneous worker attire: (1) Illumination fluctuations occur when personnel move between strong light (e.g., outdoor inspection areas) and shadow (e.g., equipment room corners), causing significant variations in pedestrian color features; (2) Workers wear standardized industrial uniforms (e.g., blue overalls, safety helmets), making texture and color features—on which state-of-the-art (SOTA) models (e.g., OSNet [23], PHA [78]) rely—nearly indistinguishable across individuals. These factors lead to a sharp decline in cross-camera matching accuracy for existing models.

**Severe Dataset Domain Gap:** Mainstream ReID datasets (e.g., Market1501 [21], Occlusion-Duke) are built on public scenarios, featuring high-angle cameras, diverse clothing styles, and stable lighting. Their data distribution differs drastically from industrial inspection scenarios. As shown in Table II, when existing models (e.g., UniHCP [74], DeepChange [75]) are directly transferred to inspection scenarios, their Rank-1 accuracy drops to below 80%, and mAP falls below 60%—a clear sign of domain shift that renders these models impractical for industrial IoT surveillance [49],[50].

To address these challenges like (1) too much background, (2) partial loss we propose a novel solution tailored for IoT surveillance systems[38],[39],[40],[41],[42],[43],[44]. In IoT surveillance settings, particularly in inspection scenarios, pedestrian detection introduces two primary issues: excessive background noise and partial pedestrian occlusion, as shown in Figure 1. These misalignments in detected images degrade feature extraction accuracy, adversely affecting ReID performance[45],[46]. Existing large-scale datasets often rely on automated detection methods, such as Deformable Parts Models (DPM), or manual annotations to address this issue [47],[48]. However, these approaches either introduce detection errors or retain alignment inconsistencies, limiting their applicability in dynamic and challenging IoT environments, such as inspections.

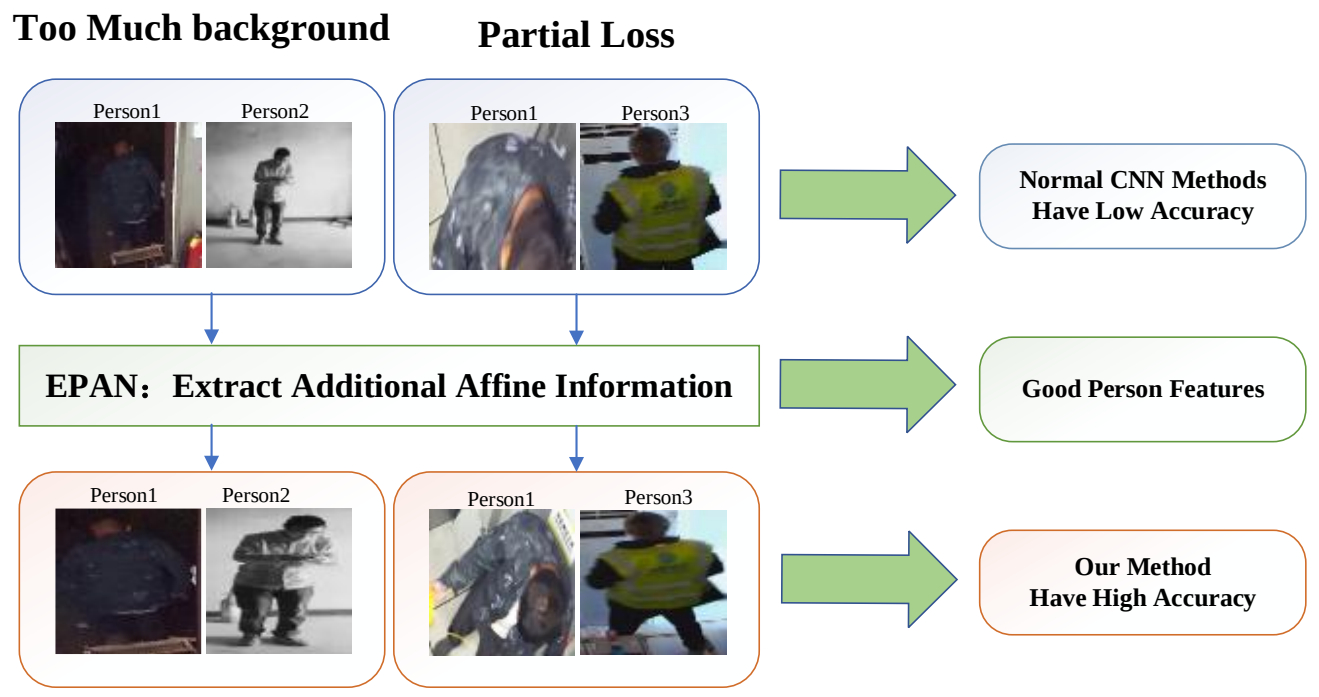

**Fig. 1.** Our EPAN effectively addresses two Issues Found in Images Processed by Detectors: a) too much background and b) Partial pedestrian loss.

To bridge this gap, we collected over 50,000 real-world pedestrian images of 1,000 unique individuals from IoT-enabled inspection staff environments over six months. This dataset was specifically curated to address the unique conditions and challenges of inspection scenarios . Building upon this, we propose the Enhanced Pedestrian Alignment Network (EPAN), a novel architecture designed to learn both global and fine-grained local features under diverse conditions, such as camera view variations and lighting changes.

EPAN introduces a dual-branch architecture: a base branch for identity prediction and an alignment branch that employs a Spatial Transformer Network (STN) to estimate affine transformation parameters for pedestrian re-alignment. By integrating alignment corrections into the feature extraction pipeline, EPAN mitigates the effects of size and position inconsistencies caused by misdetections. The training process involves three stages: identity prediction of raw inputs, affine transformation estimation, and identity prediction of the re-aligned inputs.

While conventional ReID models perform well on public datasets, their effectiveness in dynamic real-world IoT applications remains limited. EPAN is specifically trained on surveillance data from actual inspection environments. As a result, it demonstrates superior robustness and accuracy in addressing the challenges of pedestrian ReID in IoT-enabled industrial surveillance systems, making it a highly effective solution for practical deployment.

Based on the above research background and existing problems, the main contributions of this paper are summarized as follows:

### *B. Main Contributions of This Paper*

**Architectural Innovation: Dual-Branch Joint Optimization Architecture:** We propose a dual-branch architecture consisting of a "base branch (identity prediction)" and an "alignment branch (affine transformation estimation)"—the first work to embed "spatial alignment" into the ReID feature extraction pipeline. The alignment branch uses an STN module to real-time predict affine parameters, enabling "cropping (to remove redundant background)" or "zero-padding (to repair partial occlusion)" for misaligned detected images. This solves the disconnection problem of the "detection-then-recognition" pipeline in traditional single-branch models, realizing end-to-end optimization of alignment and recognition.

**Module Innovation: Enhanced Affine Network:** We design an Affine module that fuses multi-scale features: it takes Res2 (shallow local features) and Res4 (deep semantic features) from the base branch as inputs, and jointly regresses 6-dimensional affine parameters through a ResBlock and average pooling layer. Compared with traditional STN schemes that rely only on single-scale features, this module can more accurately meet alignment requirements under "viewpoint changes (e.g., low angle → eye level)" and "scale differences (e.g., close-up → long-shot)," significantly improving alignment robustness.

**Data Innovation: First Industrial IoT Inspection ReID Dataset (Inspection-Personnel):** We construct a dataset containing 1,000 unique identities and 50,000 real inspection images, covering three core scenario features: "low-angle cameras," "abrupt illumination changes," and "industrial uniforms." This fills the gap of existing datasets in the "vertical industrial field." The dataset is partitioned into training/query/test sets at a 6:1:3 ratio, providing a reliable benchmark for verifying ReID model performance in real IoT scenarios.

**Performance Innovation: SOTA-Level Practical Performance:** On the Inspection-Personnel dataset, EPAN achieves a Rank-1 accuracy of 90.09% and an mAP of 78.82% (Table III). Compared with existing SOTA methods (e.g., PHA [78]), it improves Rank-1 accuracy by 10.5% and mAP by 2.06%. Meanwhile, ResNet50 is selected as the backbone—reducing training time by 74.5% compared with ResNet101—balancing "high accuracy" and "industrial equipment deployment efficiency," and meeting the real-time requirements of IoT surveillance systems.

## II. RELATED WORK

In this section, we briefly review the related works from the following two aspects, i.e., ReID Algorithm and datasets.

### *A. Person ReID Algorithm*

Pedestrian re-identification (Re-ID) has made significant progress over the past decade [42], largely driven by advancements in deep learning techniques and the availability of large-scale annotated datasets [49],[50],[51],[52]. Early Re-ID methods primarily relied on handcrafted features such as color histograms and texture descriptors [44]. However, with the advent of convolutional neural networks (CNNs) [45], deep learning-based methods have gradually become dominant [53], offering the advantage of automatically learning discriminative features in an end-to-end manner [47], which has significantly boosted Re-ID performance [54],[55],[56].

Despite these advances [49], there remain numerous challenges in applying Re-ID methods in real-world scenarios. One major challenge is the appearance variation across different camera views [50], where the same pedestrian can appear drastically different, leading to decreased recognition accuracy [51]. In addition, issues such as occlusion, low image resolution, and cluttered backgrounds further complicate feature extraction and hinder the model's ability to distinguish individuals accurately [59],[60]. To address these challenges, various strategies have been proposed, including attention mechanisms for feature enhancement and adversarial training techniques for domain adaptation.

Among these strategies, Spatial Transformer Networks (STNs) have gained significant attention due to their ability to perform geometric transformations that aid in image alignment and feature extraction [61]. STNs reduce the impact of pose and scale variations by predicting affine transformation parameters to spatially transform the input images. Jaderberg et al. first introduced STNs into deep neural networks and demonstrated their effectiveness in various computer vision tasks [62]. Subsequently, STNs have been further applied to pedestrian re-identification to address image alignment and feature enhancement issues [63]. These algorithms include two-stage noise-tolerant, bottom-up color-independent alignment learning framework, contrastive viewpoint-aware shape learning, enhancement, integration, and expansion method, clothing status calibration, multi-view similarity aggregation and multi-level gap optimization, clothing status calibration and 3D multi-view learning network.

For the problem of unoccluded pedestrian re-identification, existing research primarily focuses on improving feature extraction and alignment strategies to enhance recognition accuracy. Zhang et al. proposed a deep network based on alignment (AlignedReID), which enhances the robustness of the model by aligning local features [64]. Sun et al. introduced the Part-based Convolutional Network (PCB), which improves the capture of local features by dividing the pedestrian image into parts for processing [65]. These methods have achieved excellent performance on public datasets, but their effectiveness in practical applications is often limited due to differences in data distribution and shooting conditions.

The two-stage noise-tolerant (TSNT) method focuses on the sensitivity of supervised person re-identification (Re-ID) approaches when dealing with label-corrupted data. In practical person Re-ID scenarios, noise labels are inevitably present in the dataset due to mislabeling or inaccurate detection. These noisy labels pose significant challenges to the effectiveness of Re-ID models. To address this issue, TSNT introduces a two-stage noise-tolerant paradigm for handling label-corrupted person Re-ID tasks [66].

The Bottom-up Color-independent Alignment Learning Framework (BCALF) proposes a color-independent alignment learning framework that decouples color-independent discrete local features and aggregates multiple key discrete features. Using color-confused images as auxiliary modalities, it performs fine-grained semantic alignment, ensuring that the smallest semantic units interact within the joint feature space, focusing on content information. BCALF designs a multi-modal encoder with shared anchors for discrete semantic alignment and aggregation of multiple key discrete features. The model is trained using color-invariant consistency constraints and inter-instance color-invariance to form a reasonable feature embedding space in color-invariant scenarios, effectively addressing the over-reliance on color in retrieval tasks [67].

Contrastive Viewpoint-aware Shape Learning (CVSL) utilizes 2D pose estimation and a shape encoder based on a Graph Attention Network (GAT) to extract local and global texture-invariant human shape cues from body poses. The Contrastive Viewpoint-aware Loss (CVL) minimizes the intra-class variance under different viewpoints and maximizes the inter-class variance within the same viewpoint, enhancing the discriminability of shape and appearance features under viewpoint variations. The Adaptive Fusion Module (AFM) integrates shape and appearance features to obtain the final representation [68].

The Enhancement, Integration, and Expansion (EIE) method is designed based on a multi-stream architecture, enhancing the model's noise awareness by using central cropping and region-based random value replacement. EIE is similar to a cascaded cross-attention mechanism, which adaptively fuses cross-stream features by studying the intrinsic interaction patterns between region and context features, enhancing the diversity and completeness of internal feature representations. EIE improves feature distinguishability and completeness through flexible keypoint matching, providing more stable and accurate features for identification [69].

The Clothing Status Calibration (MCSC) framework simulates the clothing status dynamic shift (CSDS) between meta-training and meta-testing, using a meta-optimization objective to optimize the LT-reID model, making it robust to CSDS. The MCSC framework consists of two key stages: the base model initialization stage and the meta-learning stage. In the base model initialization stage, a cross-entropy loss function is used to distinguish between different pedestrians and clothing. In the meta-learning stage, CSDS is simulated through meta-training/testing sampling and splitting strategies, introducing two loss functions—clothing status calibration loss and similarity margin ranking loss—to enhance the model's robustness to CSDS and update model parameters through the meta-optimization process [70].

The Multi-view Similarity Aggregation and Multi-level Gap Optimization (MSAMGO) method generates reliable clustering pseudo-labels by aggregating similarity matrices from multiple viewpoints, significantly reducing noise interference. MSAMGO ensures the overall quality of pseudo-labels by considering positive and negative centers of similarity at the clustering level, as well as the correlations between instance-level samples, thereby minimizing noise and jointly promoting model optimization [71].

The 3D Multi-view Learning Network (MV-3DSReID) introduces a 3D multi-view learning network, which converts 2D RGB images into 3D multi-view images using a surface random selection strategy. Four extensive 3D multi-view datasets are constructed for person Re-ID. To avoid redundancy and repetition in the 3D object rendering process, MV-3DSReID proposes a surface random selection strategy, randomly selecting two viewpoints from the six faces of a polyhedron as rendering viewpoints to generate multi-view images. MV-3DSReID includes both consistency and complementarity feature extraction, utilizing attention mechanisms and adaptive feature pooling methods to extract features from multi-view images and fuse them into a unified feature space. To fully leverage both 2D RGB images and 3D geometric information, MV-3DSReID introduces a cross-modal feature fusion strategy, combining 2D and 3D features into a unified representation space [72].

Hamza et al. proposed the S2P framework [73], which extracts a support set of source images with maximum similarity to the target data, used to maintain feature similarity during the learning process. S2P enables more robust and generalized feature learning by transferring knowledge from the teacher model to the student model through knowledge distillation (KD). Based on the similarity between source-domain and target-domain images, S2P constructs a support set for regularizing the learning process, alleviating catastrophic forgetting. By minimizing the discrepancy between source and target domains, S2P reduces domain shift, thereby enhancing stability. Existing unsupervised domain adaptation (UDA) methods are integrated into the S2P framework to improve the performance of online UDA-RID. Building on the innovative approaches mentioned above, each contributing to the advancement of person re-identification (Re-ID) in their unique ways, it is evident that the field is ripe for further exploration and enhancement. The quest for more accurate and robust Re-ID systems is relentless, as the real-world applications demand solutions that can withstand the complexities of varying environmental conditions, including camera angle variations. It is in this context that the Enhanced Pedestrian Alignment Network (EPAN) emerges as a significant contribution to the field. This novel framework is designed to tackle the challenges posed by camera angle variations in inspection staff re-identification, offering a sophisticated solution that promises to bolster the reliability and effectiveness of Re-ID systems in practical surveillance and security scenarios.

This study proposes an Enhanced Pedestrian Alignment Network (EPAN) to address the issue of camera angle variations in inspection staff re-identification. EPAN uses two convolutional neural network branches (a base branch and an alignment branch) to simultaneously perform identity prediction and affine transformation parameter estimation. The base branch extracts global features of the pedestrian image, while the alignment branch relocates the pedestrian image through affine transformation, reducing the impact of cluttered backgrounds and partial occlusions on recognition. Experimental results show that EPAN achieves robust pedestrian re-identification performance under various camera angles and conditions, demonstrating its potential for practical surveillance and security applications.

In summary, while existing pedestrian re-identification research has made significant progress on specific datasets, it still faces many challenges in practical applications. The Spatial Transformer Network, as an effective image alignment tool, provides new ideas for addressing pose and scale variation issues. For unoccluded pedestrian re-identification, the proposed EPAN effectively enhances recognition robustness and accuracy by introducing affine transformations and a multi-branch structure. These methods and improvements offer valuable insights and references for future pedestrian re-identification research.

*B. Inspection-Personnel-ReID1.0 Dataset for Vertical Field Pedestrian Re-Identification*

Pedestrian re-identification (ReID) is a critical task in surveillance and security systems, yet its performance heavily relies on domain-specific datasets. While existing datasets such as Market1501 and Occlusion-Duke provide valuable benchmarks for general scenarios, they exhibit limitations when applied to specialized vertical fields like industrial inspection personnel monitoring. This paper introduces **Inspection-Personnel-ReID1.0**, a novel dataset designed to address the unique challenges of re-identifying workers in confined inspection environments, such as equipment rooms or staff

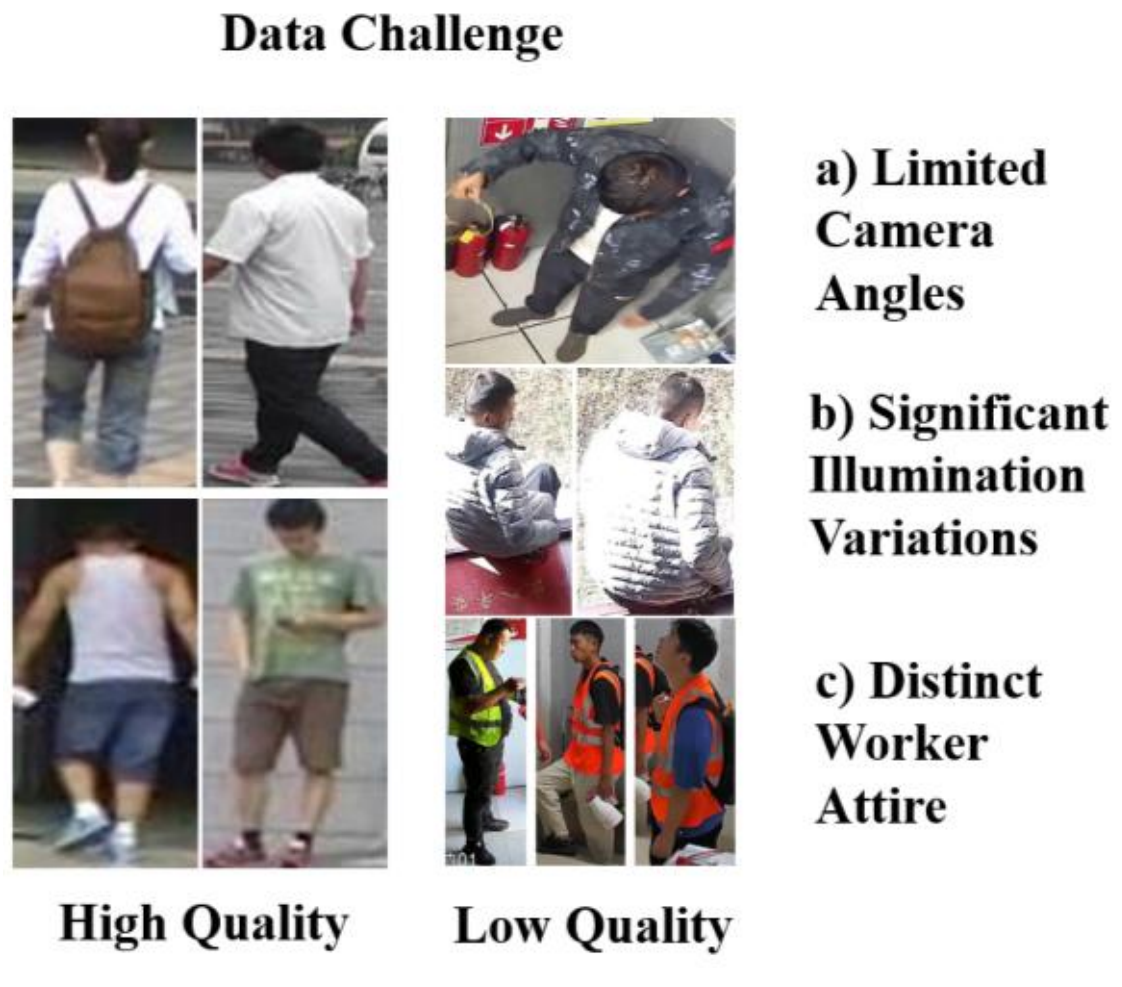

cabins.

**Fig. 2.** Three types of problems with the surveillance video frame data in the inspection staff room: a)Limited camera angles, b) Significant illumination variations and c)Distinct worker attire

**Limitations of Existing Datasets** As illustrated in Figure 2, three domain-specific challenges undermine the applicability of mainstream datasets:

1. **Low camera angles**: Surveillance cameras in inspection cabins are typically positioned below human height, contrasting with the elevated viewpoints in Market1501 and Occlusion-Duke.
2. **Severe illumination variations**: Rapid brightness fluctuations occur as personnel move through compact, artificially lit spaces.
3. **Distinctive worker attire**: Industrial uniforms exhibit homogeneous color/texture patterns distinct from public pedestrian clothing.

**Dataset Construction and Features** The Inspection-Personnel-ReID1.0 dataset was curated from surveillance footage provided by a Chinese tower company monitoring critical power infrastructure. Key construction steps include:

1. **Video processing**: Raw 1080p video streams were sampled at 10-second intervals to ensure temporal diversity while minimizing redundancy.
2. **ID annotation**: 11,091 initial IDs were filtered to exclude transient individuals (IDs with <3 occurrences), yielding 3,371 validated worker identities.
3. **Bounding box extraction**: Manually verified pedestrian regions were cropped into 24190 images (262MB total), formatted to align with Market1501 standards for compatibility.

**Dataset Partition**

- **Training set**: 14,485 images (2022 IDs, 60% of total IDs)
- **Query set**: 2,008 images (remaining IDs, 1-2 samples per ID)
- **Test set**: 7,697 images (remaining 1,349 IDs)

**Unique Advantages**

1. **Domain specificity**: Captures low-angle perspectives and industrial uniform characteristics absent in public datasets.
2. **Temporal sampling**: 10-second frame intervals mitigate pose redundancy while preserving natural movement patterns.
3. **Lighting realism**: Contains authentic illumination shifts from enclosed environments.

This dataset enables training of ReID models robust to the operational constraints of industrial monitoring systems, particularly where storage limitations necessitate efficient video data utilization. Future work will explore its potential for cross-domain adaptation and lightweight model development.

## III. Enhanced Pedestrian Alignment Network

This section will introduce our EPAN algorithm structure shown in Figure 3 in detail, including Affine Network and ResNet50, which are used in this article to automatically extract alignment information and basic features of inspectors respectively. This section will introduce our EPAN algorithm structure shown in Figure 2 in detail, including Affine Network and ResNet50, which are used in this article to automatically extract alignment information and basic features of inspectors respectively.

### *A. Alignment feature extraction*

The EPAN algorithm structure includes two modules: a traditional convolutional neural network (CNN) as the backbone network, such as ResNet50, and an affine convolutional neural network module (Affine Module) for extracting aligned features. The proposed architecture cleverly utilizes two convolutional branches and an affine estimation branch to simultaneously address these design constraints. The two main convolutional branches are referred to as the base branch and the alignment branch.

The data dimensions are typically random ($C \times H \times W$, where C is the channel number and $H \times W$ is the spatial size).To ensure the model's robustness, the input images are resized to ($3 \times 224 \times 224$). These two branches are employed to extract identity features from the training images. The base branch not only learns to distinguish images from different identities but also encodes the appearance of the images and provides spatial location information, as illustrated in Figure 3.

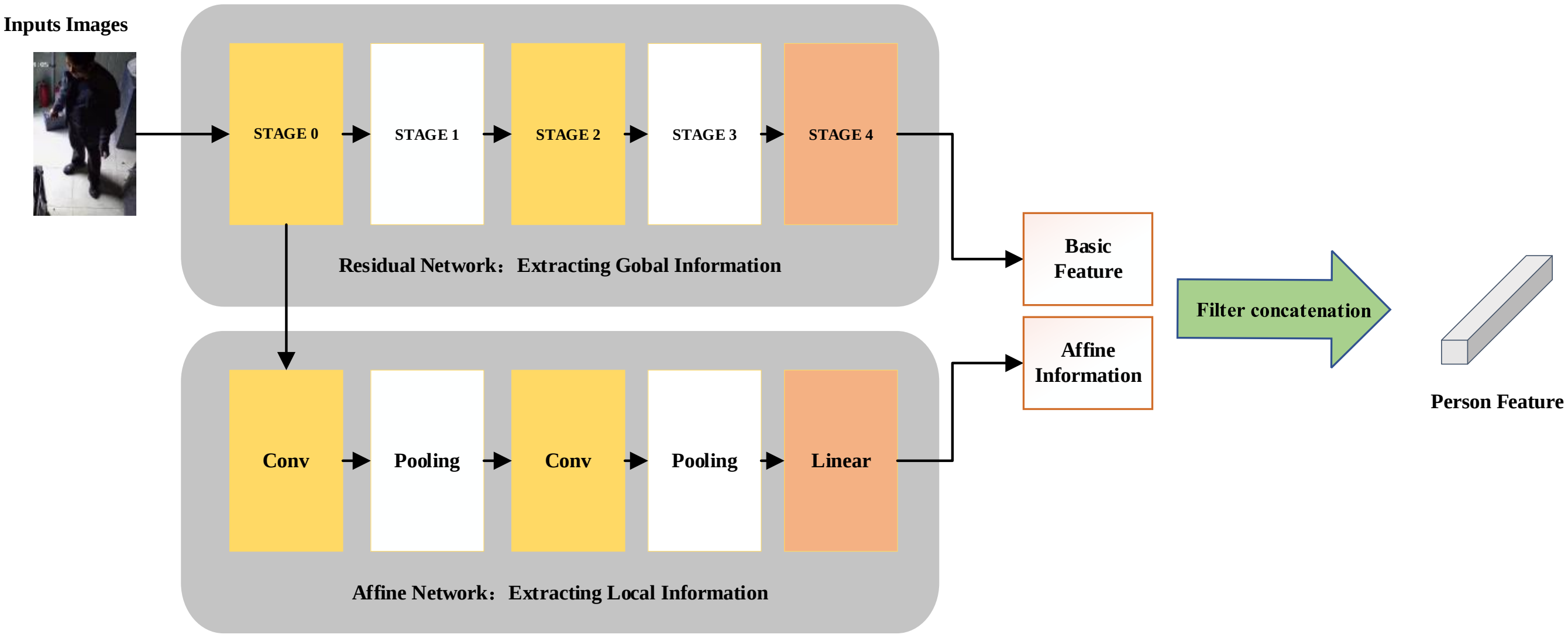

**Fig. 3**. The Affine Network and ResNet50 are respectively used to automatically extract alignment information and basic features of inspection staff personnel.

The base branch is built on a fine-tuned ResNet50. The affine branch is based on the 0th block of the base branch. The global and local features output by these two branches collectively represent pedestrian features for re-identification. To address issues of excessive background and partial occlusion, this study predicts pedestrian locations and performs spatial transformations. When there is excessive background, a cropping strategy is employed; when partial occlusion occurs, padding is applied to the image borders. Both strategies require finding the parameters of the affine transformation, which is accomplished by the affine branch.

To solve the problems of excessive background and partial occlusion, this study predicts pedestrian positions and applies spatial transformations. In the case of excessive background, a cropping strategy is used; for partial occlusion, zero-padding is applied to the image borders. Both strategies necessitate determining the parameters for the affine transformation, which are estimated by the affine estimation branch. The affine estimation branch includes a bilinear sampler and a grid network, the latter consisting of a ResBlock and an average pooling layer. Through the grid network, the Res4 feature map regresses to a set of six-dimensional transformation parameters θ, which are used to generate the image grid.

The experimental approach is straightforward, encompassing two CNN classification networks (the base branch and the alignment branch, marked in blue) and a mapping prediction network (affine estimation, marked in orange). If the target coordinates (xt, yt) are close to (m, n), the bilinear sampling method adds pixel values at the (xs, ys) position on the input feature map. The alignment branch shares a similar convolutional network but processes the aligned feature maps generated by the affine estimation branch.

*B. Affine Network*

To address the issues of excessive background and partial occlusion, the key is to predict the pedestrian's location and perform the corresponding spatial transformations. A cropping strategy is used when there is excessive background, while zero-padding is applied to the image borders in cases of partial occlusion. Both strategies require determining the parameters for the affine transformation, which is accomplished by the affine estimation branch, as illustrated in Figure 4.

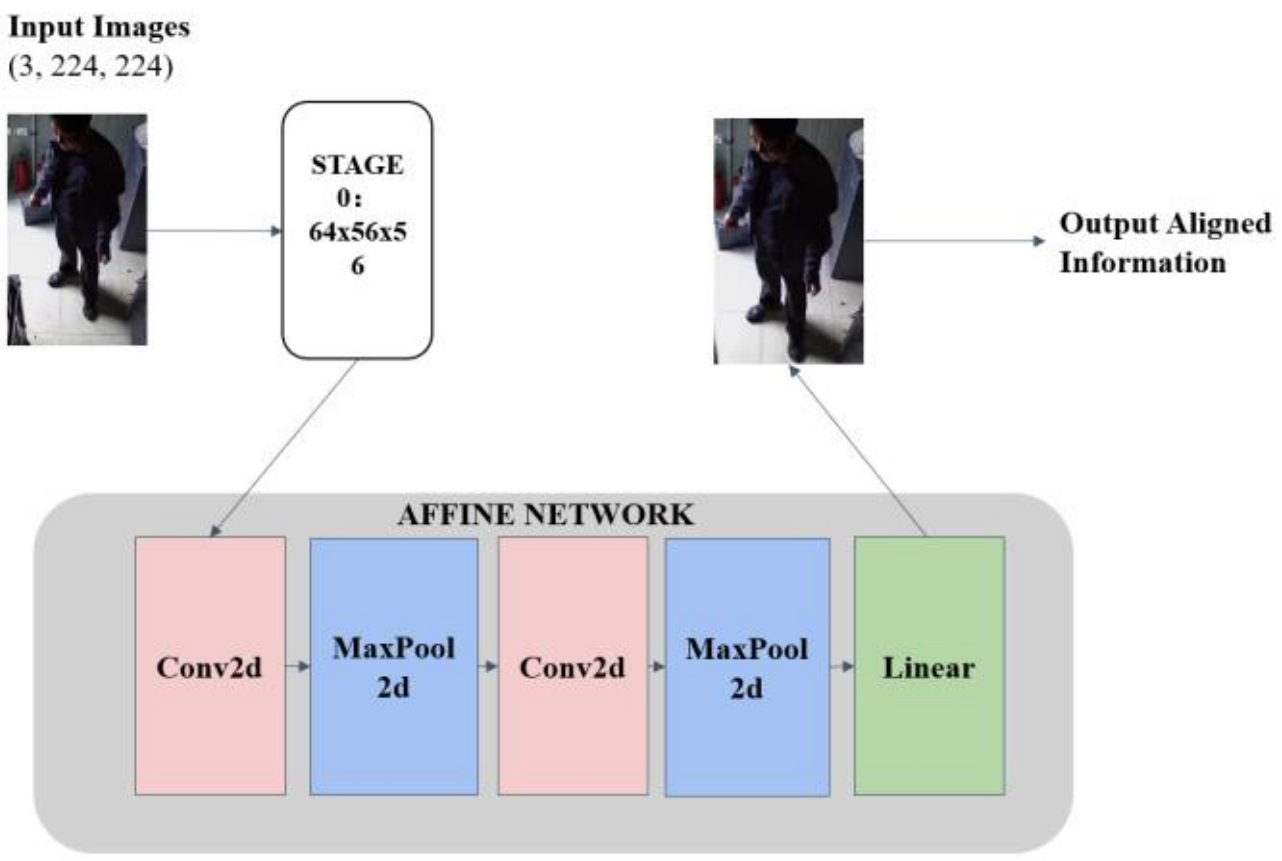

**Fig. 4.** Affine module structure for extracting alignment information of spatial affine transformations

The inputs to the affine estimation branch include two tensor activations from the base branch: 14x14x1024 and 56x56x256, corresponding to the Res4 and Res2 feature maps, respectively. The Res2 feature map contains shallow, raw image features that reflect local pattern information. In contrast, the Res4 feature map is closer to the classification layer and encodes pedestrian focal points and semantic cues that aid in identification.

The affine estimation branch consists of a bilinear sampler and a grid network, the latter comprising a ResBlock and an average pooling layer. Through the grid network, the Res4 feature map is regressed to a set of six-dimensional transformation parameters.

*C. Adaptive Loss Function*

This study considers person re-identification as a process of extracting global features, which can be understood as a classification task during the training phase. The loss function plays a crucial role in several stages of model training:

Forward Propagation: The model inputs data and performs a series of computations to produce prediction results. The predicted results are compared with the true labels to calculate the loss value.

Backward Propagation: Based on the loss value, gradients are computed for each parameter. The model parameters are then updated using the gradient descent algorithm.

Iterative Optimization: Forward and backward propagation are repeated until the loss value reaches a minimum or other stopping conditions are met.

The loss functions used in this study are as follows:

Cross Entropy with Label Smoothing: This is a cross-entropy loss function with label smoothing regularization, calculated as:

$$-\sum_{i=1}^{C}\left(y_i \cdot \log\left(\hat{y}_i + \frac{\epsilon}{C}\right)\right) \quad (1)$$

Triplet Loss: This loss function is used for person re-identification tasks and is advantageous in mining hard samples. It is calculated as:

$$\sum_{i=1}^{N} max\left(d_{pos}^2 - d_{neg}^2 + \text{margin}, 0\right) \quad (2)$$

Algorithms should be numbered and include a short title. They are set off from the text with rules above and below the title and after the last line.

## IV. Experiments

### A. Experimental Setups

The software and hardware configuration for this experiment is as follows: Lenovo Legion Y7000P 2023 model, CPU: i7-13620H, GPU: RTX4060, RAM: 40GB, Storage: 1TB SSD. The software environment includes PyCharm 2023.2.1, Python 3.8.18, and PyTorch 1.7.1.

On the basis of replication, we tested the model's performance on the inspection staff dataset, including Rank-n and mAP metrics. Extensive experimental results were used to adjust parameters and optimize the model to ensure that the final model meets the target requirements.

Experimental Design: Using the same training set, different network modules and training strategies were adjusted to determine their impact on the ReID performance.

### B. Comparative experiment

To investigate the ability of different backbone networks to extract global features, this study conducted comparative experiments using three different backbone networks on the Inspection-Personnel dataset. The three backbone networks are ResNet50, ResNeSt50, and ResNet101. By comparing the performance of these three backbone networks on the dataset, we determined which backbone network to use for subsequent experiments. The training time represents the approximate time required to train for 300 epochs under the experimental conditions. The experimental results are shown in Table I.

TABLE I
COMPARATIVE EXPERIMENTS OF DIFFERENT BACKBONE NETWORKS ON INSPECTION-PERSONNEL DATASET

| Model | Train time | Rank-1 | Rank-5 | Rank-10 | mAP |
|---|---|---|---|---|---|
| ResNet50 | **12h** | 83.45 | 89.59 | 91.52 | 57.6 |
| ResNeSt50 | 13h | 78.59 | 89.74 | 92.72 | 58.7 |
| ResNet101 | 47h | **84.92** | **90.54** | **93.82** | **60.9** |

In Table I, the backbone network with the worst performance in terms of training time and model effectiveness is ResNeSt50. The backbone network with the best performance in terms of training time is ResNet50, while ResNet101 performs the best based on model evaluation metrics. However, choosing a backbone network for pedestrian re-identification based on body posture in inspection staff requires a comprehensive consideration of both training time and model effectiveness. Although ResNet101 shows only a 1.47% improvement in Rank-1 over ResNet50, but it significantly increases training time. To meet the basic daily usage requirements of inspection staff equipment, ResNet50 is preferred as the backbone network for EPAN.

We have designed the EPAN algorithm, incorporating an affine module based on ResNet-50, which has improved the performance of the ResNet-50 model. As show in Table II our proposed algorithm, Enhanced Pedestrian Attribute Network (EPAN), demonstrates superior performance compared to existing state-of-the-art methods across all evaluated metrics. Specifically, EPAN achieves a Rank-1 accuracy of 84.92%, a Rank-5 accuracy of 90.54%, a Rank-10 accuracy of 93.82%, and a mean Average Precision (mAP) of 60.9%. These results outperform the next best method, PHA, by significant margins, particularly with a 10.5% improvement in Rank-1 accuracy and a 2.06% increase in mAP.

TABLE II
EXPLANATION OF OUR ALGORITHM'S SUPERIORITY ON INSPECTION-PERSONNEL DATASET

| Method | Rank-1 | Rank-5 | Rank-10 | mAP |
|---|---|---|---|---|
| UniHCP [74] | 76.47 | 85.7 | 88.59 | 57.58 |
| DeepChange [75] | 70.81 | 82.16 | 85.95 | 46.63 |
| HOReID [76] | 69.56 | 84.24 | 87.83 | 53.9 |
| SCSN [77] | 74.24 | 86.03 | 89.46 | 58.88 |
| OSNet [23] | 72.65 | 84.95 | 87.88 | 54.2 |
| PHA [78] | **76.86** | **88.06** | 89.53 | 58.84 |
| MSINet [79] | 72.48 | 86.48 | **90.34** | **59.74** |
| ABDNet [80] | 76.22 | 86.04 | 89.81 | 57.11 |
| Our-EPAN | **84.92** | **90.54** | **93.82** | **60.9** |

Such superiority in performance indicates that EPAN can more accurately identify and re-identify pedestrians in diverse and challenging environments, providing robust and reliable results. This level of accuracy is crucial for practical applications in security and surveillance systems, ensuring higher reliability in tracking individuals across multiple camera views and contributing to overall safety and operational efficiency.

These are figures compiled of more than one sub-figure presented side-by-side or stacked. If a multipart figure is made up of multiple figure types (one part is line art, and another is grayscale or color), the figure should meet the stricter guidelines.

## V. Algorithm Explainability

This section deeply analyzes the algorithm interpretability, including module ablation experiments, parameter selection, and Model Interpretability.

### *A. Ablation Study*

To investigate the effectiveness of the Enhanced Affine Network (AFFINE NETWORK) designed in this paper, we conducted ablation experiments on the Inspection-Personnel dataset. The experimental results are shown in Table III.

TABLE III
EPAN ablation experimental results on Inspection-Personnel

| Model | Rank-1 | Rank-5 | Rank-10 | mAP |
|---|---|---|---|---|
| S50 | 78.59 | 89.74 | 92.72 | 58.7 |
| R50-baseline | 83.45 | 89.59 | 91.52 | 57.6 |
| R101 | 89.9 | 95.54 | 96.82 | 60.9 |
| R50+Affine | 88.49 | 94.63 | 95.86 | 59.8 |
| R50+Affine+AUG | 89.59 | 94.95 | 96.27 | 60.8 |
| R50+Affine+IBN | 89.45 | 95.23 | 96.27 | 60.2 |
| **R50+Affine+AUG+IBN** | **90.09** | **95.82** | **96.86** | **65.6** |

In Table III, S50 denotes the use of ResNeSt50 as the backbone network for extracting global features, R50 represents the use of ResNet50 as the backbone network, and it is also used as a baseline model for reference in subsequent experiments. R101 indicates the use of ResNet101 as the backbone network, Affine indicates the addition of the Affine network, AUG signifies the use of data augmentation, and IBN denotes the use of instance normalization.

Based on the results of the ablation experiments, the following conclusions can be drawn:

In terms of the Rank-1 metric, using ResNet101 (R101) as the backbone network achieved the highest performance at 89.9%. In comparison, the performance using ResNeSt50 (S50) and ResNet50 (R50-baseline) was 78.59% and 83.45%, respectively.

The addition of the Affine module can enhance the model's robustness and generalization ability, making it more stable when faced with images of different angles and scales. This helps improve the model's performance on the Inspection-Personnel dataset, particularly for images with significant pose or scale variations.

### *B. Parameter analysis*

Optimizer parameters have a significant impact on the model training process. Below is an explanation of their roles:

This parameter label smooth controls whether label smoothing is enabled. Label smoothing is a regularization technique used to reduce the model's reliance on noise or uncertainty in training data, thereby improving its generalization ability. Typically, using a loss function like cross-entropy automatically adjusts model parameters, leading to overfitting of training labels. However, introducing a fixed error rate during training significantly enhances the model's generalization ability. Experimental data in Table IV demonstrates the benefits of label smoothing for the model.

TABLE IV
The help of label smoothing to the ability of model

| Setting | Rank-1 | Rank-5 | Rank-10 | mAP |
|---|---|---|---|---|
| No Lable Smoothing | 83.45 | 89.59 | 91.52 | 57.6 |
| **With Lable Smoothing** | **90.09** | **95.82** | **96.86** | **65.6** |

From the experiment results showed in Table V, it can be seen that the optimizer has a significant effect on improving the performance of the model. Adam has the advantage of adaptive learning rate and usually has good convergence effect in the same round. The momentum term in Adam algorithm can help accelerate the convergence process, especially in cases of sparse gradients or irregular surfaces. The momentum term can help SGD jump out of local minima and find the global optimal solution faster. This may be due to the label smoothing loss function reducing the sensitivity of labels, which happens to match some mistakes in the inspection staff dataset labeling that were not excluded.

TABLE V
The influence of different Optimizers on the training effect of Model

| Setting | Rank-1 | Rank-5 | Rank-10 | mAP |
|---|---|---|---|---|
| Baseline+SGD | 87.5% | 94.1% | 95.7% | 76.2% |
| Baseline+RMSProp | 79.8% | 89.3% | 92.6% | 65.3% |
| **Baseline+Adam** | **90.9%** | **96.1%** | **97.3%** | **80.6%** |

The parameter Learning-rate specifies the initial learning rate of the model, determining the step size for weight updates during training. The choice of learning rate is crucial for the training effectiveness and performance of the model. Step size: This parameter specifies the interval for learning rate decay, indicating how often the learning rate is reduced during training. gamma: This parameter specifies the decay factor for the learning rate, determining the magnitude of learning rate decay. Weight-decay: This parameter controls weight decay, a regularization technique used to mitigate model overfitting.

These optimizer parameters play a crucial role in the model training process and performance. Proper adjustment of these parameters can enhance the training effectiveness and generalization ability of the model. Reasonably adjusting these parameters can improve the training effectiveness and generalization ability of the model.

### *C. Model Interpretability*

Figure 6 shows the similarity matrix of local features:

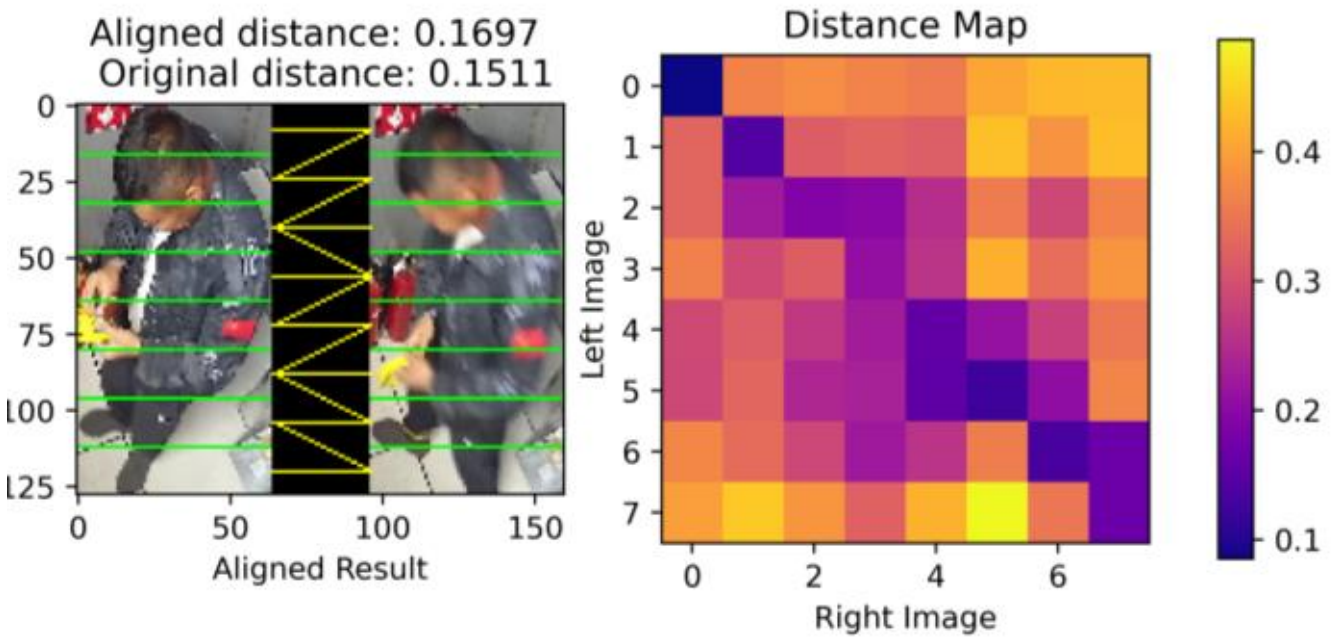

**Fig. 6.** Aligned-ReID model can be explained: block similarity matrix

As shown in Figure 6, there are two photos of the same inspection staff personnel from different angles in the left image, which are divided into 8 blocks. The Euclidean distance between each block and the other block is calculated to obtain the distance matrix of the right image. The alignment effect can be seen in the distance matrix, with aligned image blocks having smaller feature distances and misaligned image blocks having larger feature distances.

## VI. Conclusion

This paper introduces the Enhanced Pedestrian Alignment Network (EPAN), a novel approach designed to address critical challenges in pedestrian image alignment and recognition within IoT-enabled surveillance systems. By analyzing the limitations of existing methods and the issues introduced by pedestrian detection, EPAN employs a dual-branch architecture: a base branch for identity prediction and an alignment branch for estimating affine transformation parameters, effectively improving the alignment and recognition of pedestrian images.

This article proposes an Enhanced Pedestrian Alignment Network (EPAN) to address the challenges faced by pedestrian re identification (ReID) in Internet of Things (IoT) monitoring systems. Mainly completed the following three tasks:

(1) Network architecture design: EPAN architecture is proposed, which utilizes advanced alignment technology to alleviate the impact of perspective and environmental changes. The dual branch structure (the basic branch for identity prediction and the alignment branch for estimating affine transformation parameters) is used to improve the alignment and recognition ability of pedestrian images.

(2) Model optimization: Designed an enhanced affine network structure to effectively extract alignment information at different scales and perspectives; At the same time, a large number of experiments were conducted to optimize the model parameters and determine a suitable backbone network (such as selecting ResNet50 as the backbone network), which improved the performance of the model.

(3) Dataset construction and experimental validation: A large-scale real-world dataset Inspection-Personnel was constructed, containing 10000 unique individuals and 50000 images. The effectiveness of EPAN was validated through extensive experiments on this dataset. Compared with existing methods, EPAN has achieved significant improvements in both Rank-1 accuracy and Mean Accuracy (mAP). On the Inspection-Personnel dataset, the Rank-1 accuracy of EPAN reached 90.09%, and the mAP reached 78.82%. Compared with previous methods such as PHA, the Rank-1 accuracy increased by 10.5%, and the mAP increased by 2.06%.

Experimental results on the Inspection-Personnel dataset demonstrate significant performance improvements, highlighting the robustness and reliability of EPAN under real-world conditions. Notably, the parameter analysis provides insights into the impact of optimizer configurations, illustrating the loss convergence behavior and performance trends during training. These results offer an intuitive and comprehensive understanding of the model's training dynamics and its ultimate performance.

By addressing challenges such as viewpoint variations, scale inconsistencies, and low-quality image alignment, EPAN presents a practical and scalable solution for pedestrian re-identification in IoT surveillance environments. This work contributes to the vertical integration of artificial intelligence technologies in IoT-based industrial applications, such as automated personnel tracking and security monitoring. Looking forward, this research lays a solid foundation for further exploration of alignment-based models and their deployment in complex, real-world IoT systems, emphasizing the potential for broader adoption in smart surveillance and security frameworks.

## References

[1] Z. Zhang and S. Wang, “Visible thermal person Reidentification via mutual learning convolutional neural network in 6G-enabled visual Internet of Things”, *IEEE Internet of things journal*, vol. 8, no. 20, pp. 15259–15266, 2020

[2] X. Teng, C. Li, X. Li, X. Liu, and L. Lan, “TIG-CL: Teacher-Guided Individual and Group Aware Contrastive Learning for Unsupervised Person Re-Identification in Internet of Things”, *IEEE Internet of Things Journal*, 2024

[3] T. Si, F. He, P. Li, and M. Ye, “Homogeneous and heterogeneous optimization for unsupervised cross-modality

person re-identification in visual internet of things", *IEEE Internet of Things Journal*, 2023

[4] H. Sheng, Y. Zheng, W. Ke, D. Yu, X. Cheng, W. Lyu, and Z. Xiong, "Mining hard samples globally and efficiently for person reidentification", *IEEE Internet of Things Journal*, vol. 7, no. 10, pp. 9611–9622, 2020

[5] S. Liu and J. Zhang, "Local alignment deep network for infrared-visible cross-modal person reidentification in 6G-enabled Internet of Things", *IEEE Internet of Things Journal*, vol. 8, no. 20, pp. 15170–15179, 2020

[6] H. Liu, L. Xu, X. Tian, H. Peng, and D. Xia, "Visible–thermal person reidentification in visual Internet of Things with random gray data augmentation and a new pooling mechanism", *IEEE Internet of Things Journal*, vol. 10, no. 10, pp. 9022–9037, 2022

[7] S. U. Khan, I. U. Haq, N. Khan, A. Ullah, K. Muhammad, H. Chen, S. W. Baik, and V. H. C. de Albuquerque, "Efficient Person Reidentification for IoT-Assisted Cyber–Physical Systems", *IEEE Internet of Things Journal*, vol. 10, no. 21, pp. 18695–18707, 2023

[8] A. Humayed, J. Lin, F. Li, and B. Luo, "Cyber-physical systems security—A survey", *IEEE Internet of Things Journal*, vol. 4, no. 6, pp. 1802–1831, 2017

[9] H. Han, W. Ma, M. Zhou, Q. Guo, and A. Abusorrah, "A novel semi-supervised learning approach to pedestrian reidentification", *IEEE Internet of Things Journal*, vol. 8, no. 4, pp. 3042–3052, 2020

[10] X. Geng, M. Li, W. Liu, S. Zhu, H. Jiang, J. Bian, X. Fan, R. Peng, and J. Luo, "Person tracking by detection using dual visible-infrared cameras", *IEEE Internet of Things Journal*, vol. 9, no. 22, pp. 23241–23251, 2022

[11] S. Liu and J. Zhang, "Local Alignment Deep Network for Infrared-Visible Cross-Modal Person Reidentification in 6G-Enabled Internet of Things", *IEEE Internet of Things Journal*, vol. 8, no. 20, pp. 15170–15179, 2021

[12] T. Si, F. He, P. Li, and M. Ye, "Homogeneous and Heterogeneous Optimization for Unsupervised Cross-Modality Person Reidentification in Visual Internet of Things", *IEEE Internet of Things Journal*, vol. 11, no. 7, pp. 12165–12176, 2024

[13] W.-S. Zheng, J. Yan, and Y.-X. Peng, "A Versatile Framework for Multi-Scene Person Re-Identification", *IEEE Transactions on Pattern Analysis and Machine Intelligence*, pp. 1–18, 2024

[14] S. U. Khan, N. Khan, T. Hussain, and S. W. Baik, "An intelligent correlation learning system for person Re-identification", *Engineering Applications of Artificial Intelligence*, vol. 128, pp. 107213–107223, 2024

[15] M. Ye, J. Shen, G. Lin, T. Xiang, L. Shao, and S. C. H. Hoi, "Deep Learning for Person Re-Identification: A Survey and Outlook", *IEEE Transactions on Pattern Analysis and Machine Intelligence*, vol. 44, no. 6, pp. 2872–2893, 2022

[16] C. Liang, Z. Zhang, X. Zhou, B. Li, S. Zhu, and W. Hu, "Rethinking the competition between detection and reid in multi-object tracking", *IEEE Transactions on Image Processing*, vol. 31, pp. 3182–3196, 2022

[17] S. Zhang, C. Wang, and J. Peng, "ABC-Learning: Attention-Boosted Contrastive Learning for unsupervised person re-identification", *Engineering Applications of Artificial Intelligence*, vol. 133, pp. 108344–108354, 2024

[18] H. Wang, Y. Sun, and X. Bi, "Structural redundancy reduction based efficient training for lightweight person re-identification", *Information Sciences*, vol. 637, pp. 118962–118973, 2023

[19] G. Zhang, W. Lin, Arun kumar Chandran, X. Jing, and Arun kumar Chandran, "Complementary networks for person re-identification", *Information Sciences*, vol. 633, pp. 70–84, 2023

[20] D. E. Rumelhart, G. E. Hinton, and R. J. Williams, "Learning representations by back-propagating errors", *Nature*, vol. 323, no. 6088, pp. 533–536, 1986

[21] L. Zheng, L. Shen, L. Tian, S. Wang, J. Wang, and Q. Tian, "Scalable Person Re-Identification: A Benchmark", *in Proceedings of the IEEE International Conference on Computer Vision*, 2015, pp. 1116–1124.

[22] A. G. Howard, M. Zhu, B. Chen, D. Kalenichenko, W. Wang, T. Weyand, M. Andreetto, and H. Adam, "MobileNets: Efficient Convolutional Neural Networks for Mobile Vision Applications", 2017, *arXiv:1704.04861*.

[23] K. Zhou, Y. Yang, A. Cavallaro, and T. Xiang, "Omni-Scale Feature Learning for Person Re-Identification", *in Proceedings of the IEEE/CVF International Conference on Computer Vision*, 2019, pp. 3702–3712.

[24] X. Zhang, X. Zhou, M. Lin, and J. Sun, "ShuffleNet: An Extremely Efficient Convolutional Neural Network for Mobile Devices", *in Proceedings of the IEEE Conference on Computer Vision and Pattern Recognition*, 2018, pp. 6848–6856.

[25] K. He, X. Zhang, S. Ren, and J. Sun, "Deep Residual Learning for Image Recognition", *in Proceedings of the IEEE Conference on Computer Vision and Pattern Recognition*, 2016, pp. 770–778.

[26] K. Simonyan and A. Zisserman, "Very Deep Convolutional Networks for Large-Scale Image Recognition", 2015, *arXiv:1409.1556*.

[27] H. Ma, C. Zhang, Y. Zhang, Z. Li, Z. Wang, and C. Wei, "A review on video person re-identification based on deep learning", *Neurocomputing*, vol. 609, p. 128479, 2024

[28] E. Ning, C. Wang, H. Zhang, X. Ning, and P. Tiwari, "Occluded person re-identification with deep learning: A survey and perspectives", *Expert Systems with Applications*, vol. 239, p. 122419, 2024

[29] P. K. Sarker, Q. Zhao, and Md. K. Uddin, "Transformer-Based Person Re-Identification: A Comprehensive Review", *IEEE Transactions on Intelligent Vehicles*, vol. 9, no. 7, pp. 5222–5239, 2024

[30] S. U. Khan, N. Khan, T. Hussain, and S. W. Baik, "An intelligent correlation learning system for person Re-identification", *Engineering Applications of Artificial Intelligence*, vol. 128, p. 107213, 2024

[31] V. D. Nguyen, S. Mirza, A. Zakeri, A. Gupta, K. Khaldi, R. Aloui, P. Mantini, S. K. Shah, and F. Merchant, "Tackling Domain Shifts in Person Re-Identification: A Survey and Analysis", *in Proceedings of the IEEE/CVF Conference on Computer Vision and Pattern Recognition*, 2024, pp. 4149–4159.

[32] F. Zhang, H. Ma, J. Zhu, A. Hamdulla, and B. Zhu, “FRCE: Transformer-based feature reconstruction and cross-enhancement for occluded person re-identification”, *Expert Systems with Applications*, vol. 258, p. 125110, 2024
[33] K. Khaldi, V. D. Nguyen, P. Mantini, and S. Shah, “Unsupervised Person Re-Identification in Aerial Imagery”, *in Proceedings of the IEEE/CVF Winter Conference on Applications of Computer Vision*, 2024, pp. 260–269.
[34] S. Bai, H. Chang, and B. Ma, “Incorporating texture and silhouette for video-based person re-identification”, *Pattern Recognition*, vol. 156, p. 110759, 2024
[35] Y. Qin, Y. Chen, D. Peng, X. Peng, J. T. Zhou, and P. Hu, “Noisy-Correspondence Learning for Text-to-Image Person Re-identification”, *in Proceedings of the IEEE/CVF Conference on Computer Vision and Pattern Recognition*, 2024, pp. 27197–27206.
[36] M. Ye, W. Shen, J. Zhang, Y. Yang, and B. Du, “SecureReID: Privacy-Preserving Anonymization for Person Re-Identification”, *IEEE Transactions on Information Forensics and Security*, vol. 19, pp. 2840–2853, 2024
[37] B. Yang, J. Chen, and M. Ye, “Shallow-Deep Collaborative Learning for Unsupervised Visible-Infrared Person Re-Identification”, *in Proceedings of the IEEE/CVF Conference on Computer Vision and Pattern Recognition*, 2024, pp. 16870–16879.
[38] X. Zhu, X. Yao, J. Zhang, M. Zhu, L. You, X. Yang, J. Zhang, H. Zhao, and D. Zeng, “TMSDNet: Transformer with multi-scale dense network for single and multi-view 3D reconstruction”, *Computer Animation and Virtual Worlds*, vol. 35, no. 1, p. e2201, 2024
[39] Z. Xiao, Y. Chen, X. Zhou, M. He, L. Liu, F. Yu, and M. Jiang, “Human action recognition in immersive virtual reality based on multi-scale spatio-temporal attention network”, *Computer Animation and Virtual Worlds*, vol. 35, no. 5, p. e2293, 2024
[40] X. Lu, X. Xie, C. Ye, H. Xing, Z. Liu, and C. Cai, “A lightweight generative adversarial network for single image super-resolution”, *The Visual Computer*, vol. 40, no. 1, pp. 41–52, 2024
[41] H. Lu, Z. Liu, X. Pan, R. Lan, and W. Wang, “Enhancing infrared images via multi-resolution contrast stretching and adaptive multi-scale detail boosting”, *The Visual Computer*, vol. 40, no. 1, pp. 53–71, 2024
[42] N. Jiang, B. Sheng, P. Li, and T.-Y. Lee, “PhotoHelper: Portrait Photographing Guidance Via Deep Feature Retrieval and Fusion”, *IEEE Transactions on Multimedia*, vol. 25, pp. 2226–2238, 2023
[43] Y. Zhao, H. Zhang, P. Lu, P. Li, E. Wu, and B. Sheng, “DSD-MatchingNet: Deformable sparse-to-dense feature matching for learning accurate correspondences”, *Virtual Reality & Intelligent Hardware*, vol. 4, no. 5, pp. 432–443, 2022
[44] X. Li, L. Bi, J. Kim, T. Li, P. Li, Y. Tian, B. Sheng, and D. Feng, “Malocclusion Treatment Planning via PointNet Based Spatial Transformation Network”, 2020, pp. 105–114.
[45] Y.-C. Chen, X. Zhu, W.-S. Zheng, and J.-H. Lai, “Person re-identification by camera correlation aware feature augmentation”, *IEEE transactions on pattern analysis and machine intelligence*, vol. 40, no. 2, pp. 392–408, 2017
[46] M. Gou, Z. Wu, A. Rates-Borras, O. Camps, and R. J. Radke, “A systematic evaluation and benchmark for person re-identification: Features, metrics, and datasets”, *IEEE transactions on pattern analysis and machine intelligence*, vol. 41, no. 3, pp. 523–536, 2018
[47] L. Fan, T. Li, R. Fang, R. Hristov, Y. Yuan, and D. Katabi, “Learning longterm representations for person re-identification using radio signals”, 2020, pp. 10699–10709.
[48] H. Wang, X. Yuan, Y. Cai, L. Chen, and Y. Li, “V2I-CARLA: A novel dataset and a method for vehicle reidentification-based V2I environment”, *IEEE Transactions on Instrumentation and Measurement*, vol. 71, pp. 1–9, 2022
[49] Z. Dou, Y. Sun, Y. Li, and S. Wang, “Proxy-based embedding alignment for RGB-infrared person re-identification”, *Tsinghua Science and Technology*, pp. 1–13, 2024
[50] S. Dou, C. Zhao, X. Jiang, S. Zhang, W.-S. Zheng, and W. Zuo, “Human co-parsing guided alignment for occluded person re-identification”, *IEEE Transactions on Image Processing*, vol. 32, pp. 458–470, 2022
[51] X. Fang, Y. Yang, and Y. Fu, “Visible-Infrared Person Re-Identification via Semantic Alignment and Affinity Inference”, *in Proceedings of the IEEE/CVF International Conference on Computer Vision*, 2023, pp. 11270–11279.
[52] J. Miao, Y. Wu, P. Liu, Y. Ding, and Y. Yang, “Pose-Guided Feature Alignment for Occluded Person Re-Identification”, *in Proceedings of the IEEE/CVF International Conference on Computer Vision*, 2019, pp. 542–551.
[53] Z. Cui, J. Zhou, Y. Peng, S. Zhang, and Y. Wang, “Dcr-reid: Deep component reconstruction for cloth-changing person re-identification”, *IEEE Transactions on Circuits and Systems for Video Technology*, vol. 33, no. 8, pp. 4415–4428, 2023
[54] H. Luo, Y. Gu, X. Liao, S. Lai, and W. Jiang, “Bag of Tricks and a Strong Baseline for Deep Person Re-Identification”, *in 2019 IEEE/CVF Conference on Computer Vision and Pattern Recognition Workshops (CVPRW)*, 2019, pp. 1487–1495.
[55] Y. Chen, S. Zhang, F. Liu, C. Wu, K. Guo, and Z. Qi, “DVHN: A deep hashing framework for large-scale vehicle re-identification”, *IEEE Transactions on Intelligent Transportation Systems*, vol. 24, no. 9, pp. 9268–9280, 2023
[56] J. Meng, W.-S. Zheng, J.-H. Lai, and L. Wang, “Deep graph metric learning for weakly supervised person re-identification”, *IEEE Transactions on Pattern Analysis and Machine Intelligence*, vol. 44, no. 10, pp. 6074–6093, 2021
[57] F. Liu, M. Kim, Z. Gu, A. Jain, and X. Liu, “Learning Clothing and Pose Invariant 3D Shape Representation for Long-Term Person Re-Identification”, *in Proceedings of the IEEE/CVF International Conference on Computer Vision*, 2023, pp. 19617–19626.
[58] H. Rao, S. Wang, X. Hu, M. Tan, Y. Guo, J. Cheng, X. Liu, and B. Hu, “A Self-Supervised Gait Encoding Approach With Locality-Awareness for 3D Skeleton Based Person Re-Identification”, *IEEE Transactions on Pattern Analysis and Machine Intelligence*, vol. 44, no. 10, pp. 6649–6666, 2022
[59] G. Bekoulis, J. Deleu, T. Demeester, and C. Develder, “Joint entity recognition and relation extraction as a

multi-head selection problem", *Expert Systems with Applications*, vol. 114, pp. 34–45, 2018

[60] R. Wei, J. Gu, S. He, and W. Jiang, "Transformer-based domain-specific representation for unsupervised domain adaptive vehicle re-identification", *IEEE Transactions on Intelligent Transportation Systems*, vol. 24, no. 3, pp. 2935–2946, 2022

[61] B. Sun and K. Saenko, "Deep CORAL: Correlation Alignment for Deep Domain Adaptation", *Computer Vision – ECCV 2016 Workshops*, vol. 9915. pp. 443–450, 2016.

[62] M. Jaderberg, K. Simonyan, A. Zisserman, and koray kavukcuoglu, "Spatial Transformer Networks", 2015, vol. 28.

[63] Z. Zheng, L. Zheng, and Y. Yang, "Pedestrian Alignment Network for Large-scale Person Re-identification", *IEEE Trans. Circuits Syst. Video Technol.*, vol. 29, no. 10, pp. 3037–3045, 2019

[64] X. Zhang, H. Luo, X. Fan, W. Xiang, Y. Sun, Q. Xiao, W. Jiang, C. Zhang, and J. Sun, "AlignedReID: Surpassing Human-Level Performance in Person Re-Identification", 2018, *arXiv:1711.08184*.

[65] Y. Sun, L. Zheng, Y. Yang, Q. Tian, and S. Wang, "Beyond Part Models: Person Retrieval with Refined Part Pooling (and A Strong Convolutional Baseline)", *in Proceedings of the European Conference on Computer Vision*, 2018, pp. 480–496.

[66] M. Liu, F. Wang, X. Wang, Y. Wang, and A. K. Roy-Chowdhury, "A Two-Stage Noise-Tolerant Paradigm for Label Corrupted Person Re-Identification", *IEEE Transactions on Pattern Analysis and Machine Intelligence*, vol. 46, no. 7, pp. 4944–4956, 2024

[67] G. Du, H. Zhu, and L. Zhang, "Bottom-up color-independent alignment learning for text–image person re-identification", *Engineering Applications of Artificial Intelligence*, vol. 138, p. 109421, 2024

[68] V. D. Nguyen, K. Khaldi, D. Nguyen, P. Mantini, and S. Shah, "Contrastive Viewpoint-Aware Shape Learning for Long-Term Person Re-Identification", *in Proceedings of the IEEE/CVF Winter Conference on Applications of Computer Vision*, 2024, pp. 1041–1049.

[69] E. Ning, Y. Wang, C. Wang, H. Zhang, and X. Ning, "Enhancement, integration, expansion: Activating representation of detailed features for occluded person re-identification", *Neural Networks*, vol. 169, pp. 532–541, 2024

[70] Y. Huang, Q. Wu, Z. Zhang, C. Shan, Y. Huang, Y. Zhong, and L. Wang, "Meta Clothing Status Calibration for Long-Term Person Re-Identification", *IEEE Transactions on Image Processing*, vol. 33, pp. 2334–2346, 2024

[71] T. Liu, S. Cheng, and A. Du, "Multi-view similarity aggregation and multi-level gap optimization for unsupervised person re-identification", *Expert Systems with Applications*, vol. 256, p. 124924, 2024

[72] Z. Yu, L. Li, J. Xie, C. Wang, W. Li, and X. Ning, "Pedestrian 3D Shape Understanding for Person Re-Identification via Multi-View Learning", *IEEE Transactions on Circuits and Systems for Video Technology*, vol. 34, no. 7, pp. 5589–5602, 2024

[73] H. Rami, J. H. Giraldo, N. Winckler, and S. Lathuilière, "Source-Guided Similarity Preservation for Online Person Re-Identification", *in Proceedings of the IEEE/CVF Winter Conference on Applications of Computer Vision*, 2024, pp. 1711–1720.

[74] Y. Ci, Y. Wang, M. Chen, S. Tang, L. Bai, F. Zhu, R. Zhao, F. Yu, D. Qi, and W. Ouyang, "UniHCP: A Unified Model for Human-Centric Perceptions", *in Proceedings of the IEEE/CVF Conference on Computer Vision and Pattern Recognition*, 2023, pp. 17840–17852.

[75] P. Xu and X. Zhu, "DeepChange: A Long-Term Person Re-Identification Benchmark with Clothes Change", *in Proceedings of the IEEE/CVF International Conference on Computer Vision*, 2023, pp. 11196–11205.

[76] G. Wang, S. Yang, H. Liu, Z. Wang, Y. Yang, S. Wang, G. Yu, E. Zhou, and J. Sun, "High-Order Information Matters: Learning Relation and Topology for Occluded Person Re-Identification", *in Proceedings of the IEEE/CVF Conference on Computer Vision and Pattern Recognition*, 2020, pp. 6449–6458.

[77] X. Chen, C. Fu, Y. Zhao, F. Zheng, J. Song, R. Ji, and Y. Yang, "Salience-Guided Cascaded Suppression Network for Person Re-Identification", *in Proceedings of the IEEE/CVF Conference on Computer Vision and Pattern Recognition*, 2020, pp. 3300–3310.

[78] G. Zhang, Y. Zhang, T. Zhang, B. Li, and S. Pu, "PHA: Patch-Wise High-Frequency Augmentation for Transformer-Based Person Re-Identification", *in Proceedings of the IEEE/CVF Conference on Computer Vision and Pattern Recognition*, 2023, pp. 14133–14142.

[79] J. Gu, K. Wang, H. Luo, C. Chen, W. Jiang, Y. Fang, S. Zhang, Y. You, and J. Zhao, "MSINet: Twins Contrastive Search of Multi-Scale Interaction for Object ReID", *in Proceedings of the IEEE/CVF Conference on Computer Vision and Pattern Recognition*, 2023, pp. 19243–19253.

[80] T. Chen, S. Ding, J. Xie, Y. Yuan, W. Chen, Y. Yang, Z. Ren, and Z. Wang, "ABD-Net: Attentive but Diverse Person Re-Identification", *in Proceedings of the IEEE/CVF International Conference on Computer Vision*, 2019, pp. 8351–8361.